# Analyzing the Effects of Reasoning Types on Cross-Lingual Transfer Performance

**Karthikeyan K**[*]
Duke University
karthikeyan.k@duke.edu

**Aalok Sathe**[*]
Massachusetts Institute of Technology
asathe@mit.edu

**Somak Aditya** and **Monojit Choudhury**
Microsoft Research India
{t-soadit,monojitc}@microsoft.com

## Abstract

Multilingual language models achieve impressive zero-shot accuracies in many languages in complex tasks such as Natural Language Inference (NLI). Examples in NLI (and equivalent complex tasks) often pertain to various types of sub-tasks, requiring different kinds of reasoning. Certain types of reasoning have proven to be more difficult to learn in a monolingual context, and in the crosslingual context, similar observations may shed light on zero-shot transfer efficiency and few-shot sample selection. Hence, to investigate the effects of types of reasoning on transfer performance, we propose a category-annotated multilingual NLI dataset and discuss the challenges to scale monolingual annotations to multiple languages. We statistically observe interesting effects that the confluence of reasoning types and language similarities have on transfer performance.

## 1 Introduction

Recent work has shown that masked language models such as XLM (Lample and Conneau, 2019), mBERT (Devlin et al., 2019) and XLM-RoBERTa (Conneau et al., 2019) can achieve efficient crosslingual transfer, in both zero-shot and few-shot settings. Such results have motivated researchers (K et al., 2020; Artetxe et al., 2020; Conneau et al., 2020) to investigate the underlying influencing factors behind transfer efficiency along different dimensions such as model capacity, language similarity and learning objective. However, such transfer efficiency is measured for natural language understanding (NLU) tasks which often cover a wide range of linguistic phenomena. Natural Language Inference is one such representative task which is known to require different types of reasoning and linguistic phenomena. In the monolingual context, recently, authors in Joshi et al. (2020a) extended previous work (Richardson et al., 2020; Salvatore et al., 2019) and listed a comprehensive list of types of reasoning capabilities required to solve the NLI examples in large public NLI datasets. Authors show that NLI requires a mix of 15 different types of reasoning capabilities (TAXINLI), categorized broadly into LANGUAGE, LOGIC and KNOWLEDGE. They observe that both BERT (Devlin et al., 2019) and RoBERTa (Liu et al., 2019) perform poorly in reasoning tasks such as *causal* and *coreference*, whereas they pick up *negation* easily; and observe that some categories are intrinsically harder for these models (both Transformers-based and pre-Transformers LSTM-based) to understand. Our central hypothesis of this work is, *crosslingual transfer gap* (Hu et al., 2020) and *few-shot performance* may depend on the type of reasoning required. To this end, we propose multilingual extension of the TAXINLI dataset. Our zero-shot analysis strongly suggests that reasoning types play a critical role in transfer performance. In few-shot, some categories (such as Negation), show steady improvement in accuracy along-with increased cross-lingual alignment. For others (syntactic, logic) improvement varies across language. For causal, few-shot models fail to generalize well on a diagnostic gold set.

## 2 TaxiXNLI: Towards Multilingual Extension of TaxiNLI

In monolingual settings, various authors (Nie et al., 2019; Wang et al., 2018; Ribeiro et al., 2020) have proposed a shift from tracking end-to-end task accuracy to explicit categorizations of fundamental capabilities and to track such individual category-wise errors. While some of the proposed categories seem to capture relevant reasoning capabilities, they are often incomplete as they are tuned towards analyzing errors. Joshi et al. (2020a) recently proposed a categorization of the reasoning tasks involved in NLI. It takes inspiration from earlier literature in linguistics (Wittgenstein, 1922) and logic (Sowa, 2010). According to Wittgen-stein (1922), humans require a combination (or a sequence) of lexical, syntactic, semantic and pragmatic reasoning to gauge *meaning* from language (or *form*). The TAXINLI categorization scheme, which defines three broad groups of reasoning LINGUISTIC, LOGICAL and KNOWLEDGE, aligns with our philosophy. Additionally, the proposed categorization is also useful because it retains the categories that are relevant for public NLI datasets, and the granularity is well-defined. Further, the availability of the annotated dataset makes it a perfect choice for the current study.

---
[*]Work done while at Microsoft Research India.

We introduce the categories briefly. For detailed definitions and examples, we refer the readers to Joshi et al. (2020a). LINGUISTIC represents NLI examples where inference process is internal to the provided text; further divided into three categories `lexical`, `syntactic` and `factivity`. LOGICAL denotes examples where inference may involve processes external to text, and grouped under *Connectives*, and *Deduction*. *Connectives* involve categories such as `Negation`, `boolean`, `quantifiers`, `conditionals` and `comparatives`. Lastly, *Deduction* involves different types of reasoning such as `relational`, `spatial`, `causal`, `temporal` and `coreference`. Knowledge indicates examples where external (`world`) or commonly assumed knowledge (`commonsense`) is required for inferencing.

|  | Noise | | F1 | Acc | $tr = 0, en = 1$ |
|---|---|---|---|---|---|
|  | **Partial** | **Full** |  |  |  |
| **Hindi** | 18.4 | 8.9 | 0.22 | 0.81 | 0.44 |
| **Spanish** | 38.3 | - | 0.25 | 0.69 | 0.32 |
| **Swahili** | 39.8 | 36.8 | 0.21 | 0.84 | 0.34 |

Table 1: We show the percentage of full and partial noisy samples; F1, Accuracy scores with respect to the English TAXINLI annotations for each language on the sampled set of examples. We also include percentage of category annotations where the annotated target category is zero, while the corresponding category in English is 1.

### 2.1 TaxiXNLI Dataset

**TaxiXNLI (translated)** The TAXINLI dataset (Joshi et al., 2020a) consists of 10071 premise, hypothesis (P-H) pairs drawn from the MultiNLI (MNLI) dataset (Williams et al., 2018a). Each pair is annotated with the types of reasoning (among 15 categories) that are required to make the inference. To account for the asymmetric distribution of examples across taxonomic categories, we use the hierarchical taxonomy provided by Joshi et al. (2020a) to group some of the related categories together[1]. For our crosslingual analysis, we split the English examples equally[2] into train and test set (5k/5k) by balancing across taxonomic categories through iterative stratification (Sechidis et al., 2011). Inspired by prior work (Gerz et al., 2018; Joshi et al., 2020b), we choose Spanish, French, Russian, Hindi, Arabic, Vietnamese, Chinese, Swahili, and Urdu; which collectively cover a spectrum of varying resource (high - 4, medium - 3, low - 2) and typological diversity. We use the *Bing translator* to translate both train and test set into these nine languages.

Due to the noise inherent in automatic translation, first we perform a qualitative analysis of translated examples. Following Joshi et al. (2020a), we conduct manual annotations of TAXINLI categories for a subset of translated pairs in three languages: Swahili, Hindi and Spanish, by native speakers. In addition to categories, we instruct them to indicate full, partial, and no noise translations. Among them, there were 8.95% (18.4) and 36.81% (39.8) were fully (and partially) noisy sentence-pairs in Hindi and Swahili, and Spanish had 38% partially noisy sentences. The percentage of examples where a category in English is 1, but the same in a translated language is zero are 44 (hi), 34 (sw) and 32% (es) respectively. Upon further analysis and interviews with the annotators, we discover changes (from 1 to 0) are due to 2 reasons: 1) partially noisy translations that make the premise and hypothesis somewhat unrelated; and, the TAXINLI instructions instruct annotators not to proceed with category annotations for un-related neutral examples; 2) difference in reasoning adapted by individual annotators. Our analysis (detailed in next section), aligns with the intuition from the definitions that, translation to different languages does not change the reasoning categories in the LOGICAL and KNOWLEDGE bucket.

**TaxiXNLI (diagnostic)** Motivated by the above study, we look towards the XNLI dataset (Conneau et al., 2018), that provides parallel P-H pairs in 15 languages. We sample 1.4k XNLI examples and annotate with a few selected interesting categories (Negation, Boolean, Spatial, Causal, Temporal, Knowledge). This is used as a diagnostic gold test set for our zero-shot and few-shot experiments.

---
[1]Logic: {Quantifier, Conditional, Comparative}; Deductions: {Relational, Spatial, Coreference, Temporal}; Knowledge: {World, Taxonomic}

[2]Category-wise splits are in Tab. 4.

| | Category | P (eng) | H (eng) | P (target) | H (target) |
|---|---|---|---|---|---|
| | | | Swahili | | |
| E1 | world (eng) | Middle Ages | The time period during which the Black Plague happened. | Zama za kati | Kipindi cha wakati ambapo pigo la Black lilitokea. |
| E2 | taxonomic (swa) | Carmel Man, a relation of the Neanderthal family, lived here 600,000 years ago. | Carmel Man isn't alive today. | Karmeli Man, uhusiano wa familia ya Neanderthal, aliishi hapa miaka 600,000 iliyopita. | Karmeli Man si hai leo. |
| E3 | spatial (swa) | I'm from New York. | I'm from Texas. | Mimi ni kutoka New York. | Mimi ni kutoka Texas. |
| E4 | temporal (eng) | They need individual support to attend conferences, present papers, publish their works, and keep in touch with others in their fields across the country. | They need a lot of funding to do their work effectively. | Wanahitaji msaada wa kibinafsi kuhudhuria mikutano, kuwasilisha makaratasi, kuchapisha kazi zao, na kuwasiliana na wengine katika nyanja zao nchini kote. | Wanahitaji fedha nyingi kufanya kazi zao kwa ufanisi. |
| E5 | temporal (swa) | So he says, No, from the school up, are all guerrillas. | He says everyone in this community is pacifistic. | Hivyo anasema, Hapana, kutoka shule, wote ni waasi. | Anasema kila mtu katika jamii hii ana mpango wa kutelezi. |
| E6 | negation (eng) | So it's almost like a second home. | Alas, I have never been to such a place. | Hivyo ni karibu kama nyumba ya pili. | Ole, sijawahi kuwa na sehemu hiyo. |
| E7 | negation (swa) | He argued that other extremists, who aimed at local rulers or Israel, did not go far enough. | He held the stance that the other extremists should have went further. | Alisema kwamba wengine wenye siasa kali, ambao walikusudia viongozi wa ndani au Israeli, hawakwenda mbali sana. | Yeye alishikilia msimamo kwamba wengine wenye siasa kali wanapaswa kwenda zaidi. |

Table 2: Interesting examples, where a category is marked in one language, but not in the other. The categories are mentioned in the second column, and in braces, we mention the language where the category is marked present. For example, in E1, category "world" is marked only in English. E1: a translation error, so marked neutral, E2: annotator needed to look up "Karmeli man", and understand relation with Neanderthal, E3: Swahili annotator used geographic reasoning, E4: English annotation is dubious, E5: wrong annotation in Swahili, E6: partial translation error, so neutral, E7: Swahili annotator misunderstood the sentence.

| | Category | P (eng) | H (eng) | P (target) | H (target) |
|---|---|---|---|---|---|
| | | | Hindi | | |
| E1 | world (eng) | As a result, their services may be more effective when conducted in the emergency department environment. | Their services might be more effective if they're done in the ED. | नतीजतन, आपातकालीन विभाग के माहौल में आयोजित होने पर उनकी सेवाएं अधिक प्रभावी हो सकती हैं। | यदि वे ईडी में किए जाते हैं तो उनकी सेवाएं अधिक प्रभावी हो सकती हैं। |
| E2 | world (hi) | Twenty islands have since been added, making the Cyclades the largest of the Greek island groups. | They have added just three more islands. | बीस द्वीपों के बाद से जोड़ा गया है, Cyclades ग्रीक द्वीप समूहों के सबसे बड़ा बना । | उन्होंने सिर्फ तीन और द्वीप जोड़े हैं । |
| E3 | taxonomic (eng) | Favored by the Ancient Egyptians as a source of turquoise, the Sinai was, until recently, famed for only one event but certainly an important one. | The Sinai was a source of gold for the Ancient Egyptians. | फिरोजा के एक स्रोत के रूप में प्राचीन मिस्र के इष्ट, सिनाई, हाल ही में जब तक, केवल एक घटना के लिए प्रसिद्ध है, लेकिन निश्चित रूप से एक महत्वपूर्ण एक था । | सिनाई प्राचीन मिस्रियों के लिए सोने का स्रोत था। |
| E4 | taxonomic (hi) | 'And I don't want to risk a fire fight with what appear to be horribly equal numbers.' | I don't want to get in a fight. | और मैं क्या बुरी तरह से बराबर संख्या में दिखाई देते हैं के साथ एक आग लड़ाई जोखिम नहीं करना चाहती । | मैं लड़ाई में नहीं पड़ना चाहता । |
| E5 | spatial (eng) | On the slopes of the hill you will find Edinburgh Zoo, located just behind Corstorphine Hospital. | Corstophine hospital is really far fromEdinburgh Zoo | पहाड़ी की ढलानों पर आपको एडिनबर्ग चिड़ियाघर मिलेगा, जो कॉस्टॉर्फिन अस्पताल के ठीक पीछे स्थित है। | Corstophine अस्पताल वास्तव में एडिनबर्ग चिड़ियाघर से दूर है |
| E6 | temporal (eng) | Both individuals agreed that the teleconference played no role in coordinating a response to the attacks of 9/11. | Neither party claimed that the phone meeting of the prior morning had been responsible for the actions taken in relation to the attacks on 9/11. | दोनों व्यक्तियों ने इस बात पर सहमति व्यक्त की कि टेलीकांफ्रेंस ने 9/11 के हमलों के प्रत्युत्तर के समन्वय में कोई भूमिका नहीं निभाई । | न तो पार्टी ने दावा किया कि पूर्व सुबह की फोन बैठक 9/11 पर हमलों के संबंध में की गई कार्रवाइयों के लिए जिम्मेदार रही थी । |
| E7 | temporal (hi) | By 1993, Penney was using EDI for processing 97 percent of purchase orders and 85 percent of invoices with 3,400 of its 4,000 suppliers. | Penney was using EDI in 1993 for processing much of its supply information. | 1993 तक, पेनी खरीद आदेशों के 97 प्रतिशत और 85 प्रतिशत चालानों को अपने 4,000 आपूर्तिकर्ताओं के 3,400 के साथ संसाधित करने के लिए ईडीआई का उपयोग कर रहा था। | पेनी अपनी आपूर्ति की अधिकांश जानकारी के प्रसंस्करण के लिए 1993 में ईदी का उपयोग कर रहा था। |
| E8 | negation (eng) | The handling and processing of the reagent, commonly limestone, is often done onsite, as is the treatment of the effluent as waste or processing into a saleable product (e.g. | The reagent is more often than not limestone, and is handled and processed onsite. | अभिकर्प, आमतौर पर चूना पत्थर की हैंडलिंग और प्रसंस्करण अक्सर ऑनसाइट किया जाता है, जैसा कि अपशिष्ट या प्रसंस्करण के रूप में बहिःस्राव का उपचार एक बिक्री योग्य उत्पाद (उदाहरण के लिए) में किया जाता है। | अभिवाक् त चूना पत्थर की तुलना में अधिक बार होता है, और इसे ऑनसाइट संभाला और संसाधित किया जाता ह। |
| E9 | negation (hi) | CHAPTER 6: HUMAN CAPITAL | Capital is money, not people. | अध्याय 6: मानव पूंजी | पैसा है, लोगों को नहीं। |

Table 3: Similar examples in Hindi. The categories are mentioned in the second column, and in braces, we mention the language where the category is marked present.

We will share the dataset publicly in github.com/microsoft/TaxiXNLI.

## 2.2 Automated Translation Error Analysis

To gauge the errors due to automated translation of English to different languages and inheriting the reasoning types from TAXINLI annotations, we hire native speakers in Hindi, Spanish and Swahili to re-annotate 200 carefully balanced samples. The annotators were trained similarly as described in Joshi et al. (2020a); and were additionally instructed to label noisy (partial or full) example pairs. We share the results in Table 1.

**Interviews and Further Analysis** Our goal was to investigate two questions: 1) how noisy are the translations, 2) for non-noisy translations, do a reasoning category stay the same across languages.

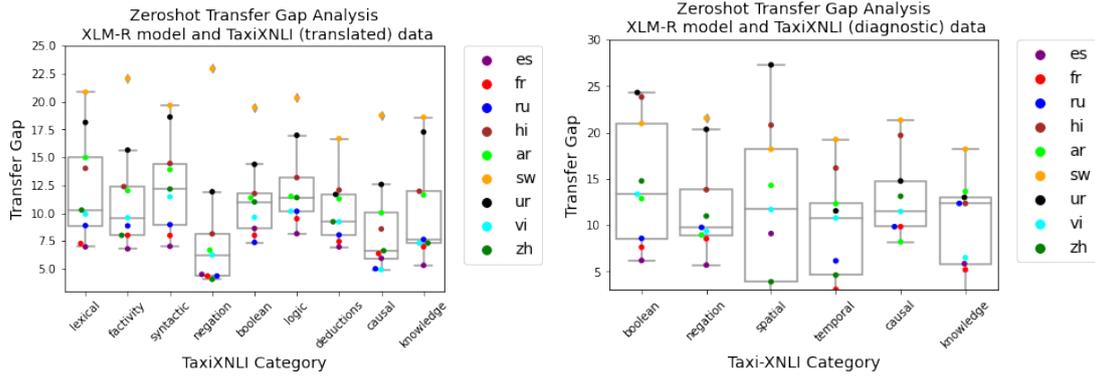

Figure 1: For each TaxiNLI category and language, we plot the transfer gap on XLM-R, for both datasets.

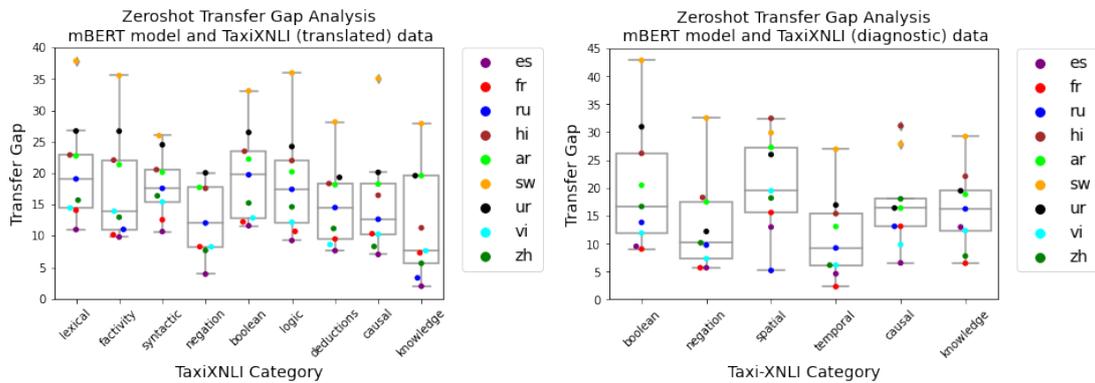

Figure 2: For each TaxiNLI category and language, we plot the transfer gap on mBERT, for both datasets.

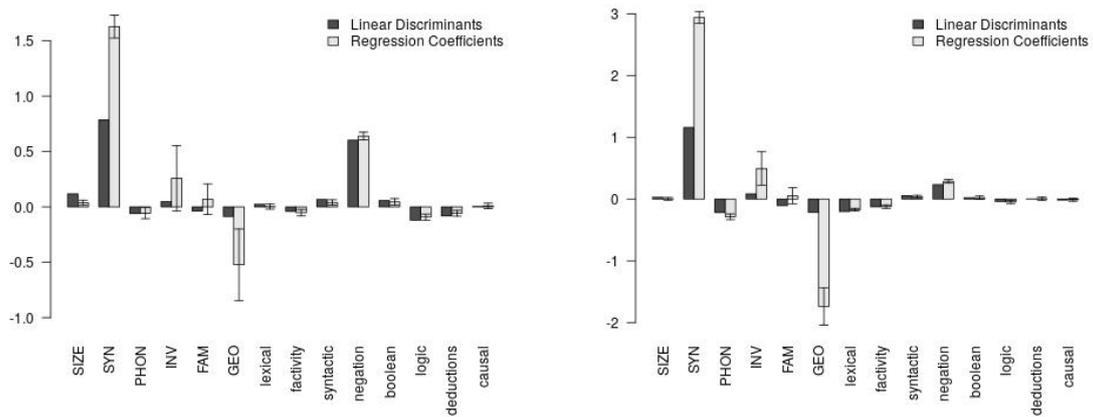

Figure 3: LDA and Regression coefficients obtained to model the NLI prediction correctness on TaxiXNLI (translated) by XLM-R (left) and M-BERT (Right), given language features and reasoning categories. (SYN/negation: ***, Logic: *, $p$-value $\leq$ 0***, 0.001**, 0.01*)

While the noise level in both Hindi and Spanish is lower, translation accuracy for low-resource lan-guages are still pretty poor. This prompted us to create the TaxiXNLI (diagnostic). We wanted to further investigate the second question, as the percentage of #tr=0,en=1 (indicating the categories which was present in English, but was not when translated to the corresponding target language) is lower than expected. After conducting an interview with the Swahili and the Hindi annotators about mismatching annotations, we found the following: 1) partially noisy translations often made the premise and hypothesis somewhat unrelated; and, based on TaxiNLI instructions annotators did not further annotate any categories; 2) TaxiNLI reasoning category annotations are subjective, as different annotators can follow different reasoning process. For example, for a *P: I am from New York. H: I am from Texas*; the Swahili translation was perfect *P: Mimi ni kutoka New York., H: Mimi ni kutoka Texas.*. But the Swahili annotator reasoned geographically, i.e., N.Y is far from Texas, so annotated it as *spatial* and the English annotator reasoned using world knowledge; 3) sometimes translation also slightly changed the word sense. For example, for the hypothesis *H: He held the stance that the other extremists should have went further.*, the translation was *Yeye alishikilia msimamo kwamba wengine wenye siasa kali wanapaswa kwenda zaidi.*, which loosely translates to "extremists should have went further in distance". More examples are in Table 2 and 3.

## 2.3 Dataset Statistics

| Category | Translation | | Diagnostic |
|---|---|---|---|
| | Train | Test | Test |
| lexical | 1362 | 1362 | - |
| factivity | 824 | 824 | - |
| syntactic | 1439 | 1438 | - |
| negation | 688 | 688 | 246 |
| boolean | 801 | 800 | 210 |
| logic | 895 | 896 | 207* |
| deductions | 1019 | 1019 | - |
| causal | 1176 | 1177 | 61 |
| knowledge | 301 | 301 | 154 |

Table 4: **Number of examples:** For each TaxiNLI categories, we report number of train and test examples in TaxiXNLI (translated) and TaxiXNLI (Diagnostic) respectively. * within logic, we only annotate spatial and temporal examples.

For both TaxiXNLI (translated) and TaxiXNLI (diagnostic), we report the number of training and test examples for each TaxiNLI categories in Table 4. In the diagnostic set, we annotate a total 1435 examples, among which there were 650 examples, where no selected category was marked.

| | lex | fact | synt | neg | bool | log | ded | caus | know |
|---|---|---|---|---|---|---|---|---|---|
| mBERT | 20.52 | 18.19 | 18.21 | 18.38 | 19.68 | 18.52 | 15.06 | 15.41 | 11.59 |
| XLM-R | 12.37 | 11.49 | 12.69 | 8.14 | 11.29 | 12.38 | 10.29 | 8.76 | 10.45 |
| Δ | 8.15 | 6.7 | 5.52 | 10.24 | 8.39 | 6.14 | 4.77 | 6.65 | 1.14 |

Table 5: **Transfer Gap and TaxiNLI categories:** For each category, we report the transfer gap averaged across languages and the reduction in the gap when we move from M-BERT to XLM-R.

## 3 Experimental Setup

For our experiments, we fine-tune Multilingual BERT and XLM-R (base)[3] on MultiNLI training (Williams et al., 2018b) data and select the best model using the MultiNLI validation set. We repeat the fine-tuning with 5 different seed values and for learning-rate in {2e-5, 3e-5}. For zero-shot experiments, we evaluate the best model on TaxiXNLI datasets for each language and taxonomic category. We measure the crosslingual transferability using transfer gap (Hu et al., 2020) which is defined as the difference between the performance on English and target language.

Whereas, for few-shot learning, for each of the target language and inference type pair, we continue to fine-tune the best MultiNLI fine-tuned XLM-R and M-BERT model using a random few TaxiXNLI (translated) training examples in that particular pair and then test the model on the same language and inference type pair. We repeat all our experiments with 5 different seeds (different examples each time) and report the average. For our analysis, we measure relative error reduction (RER = $\frac{(\text{accuracy}_N - \text{accuracy}_0)}{100 - \text{accuracy}_0}$) which provides measure of the error reduction from the initial step (zero-shot), relative to the initial step.

## 4 CrossLingual Cross-capability Insights

In this section, we discuss some important insights that we observe through the zero-shot and few-shot experiments. We highlight phenomena which are repeated across TaxiXNLI (translated) and TaxiXNLI (diagnostic) datasets.

**Is Transferability uniform across categories? Does more pre-training help?** From Figure 1

---

[3] github.com/huggingface/transformers

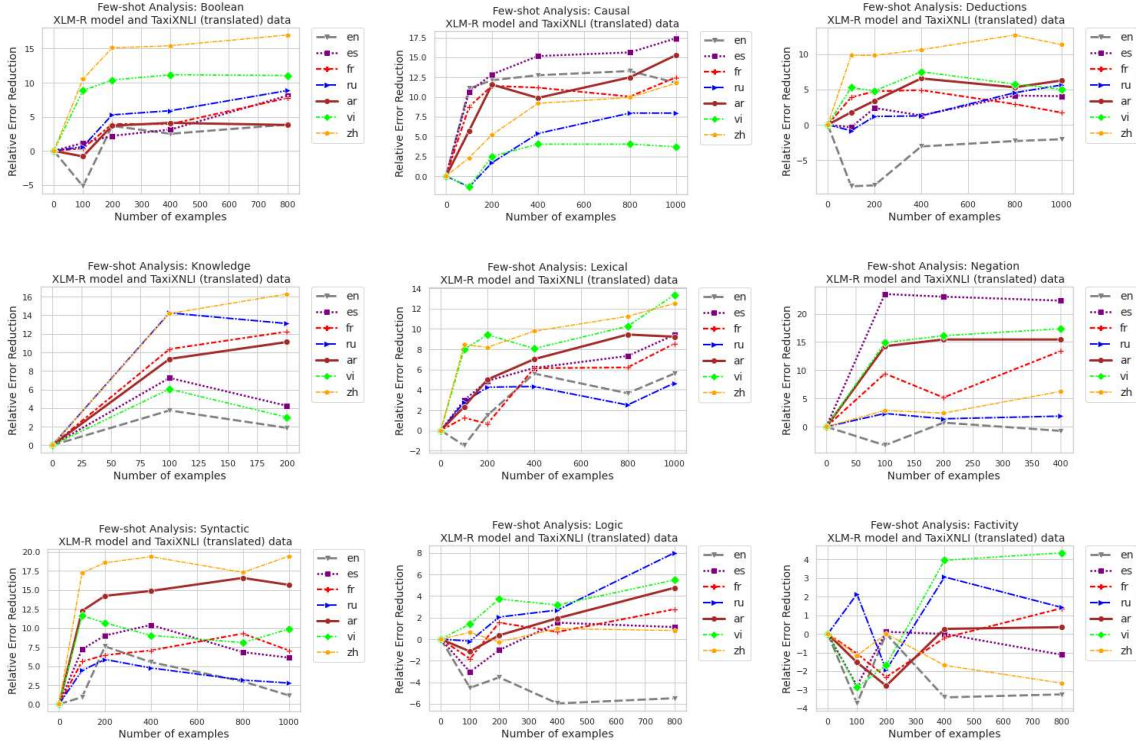

Figure 4: **Few Shot Accuracy (TAXIXNLI (translated), XLM-R):** For each TAXINLI category and language, we plot Accuracy on TAXIXNLI (translated) test, when XLM-R MultiNLI model is fine-tuned on $N$ examples in that language and category

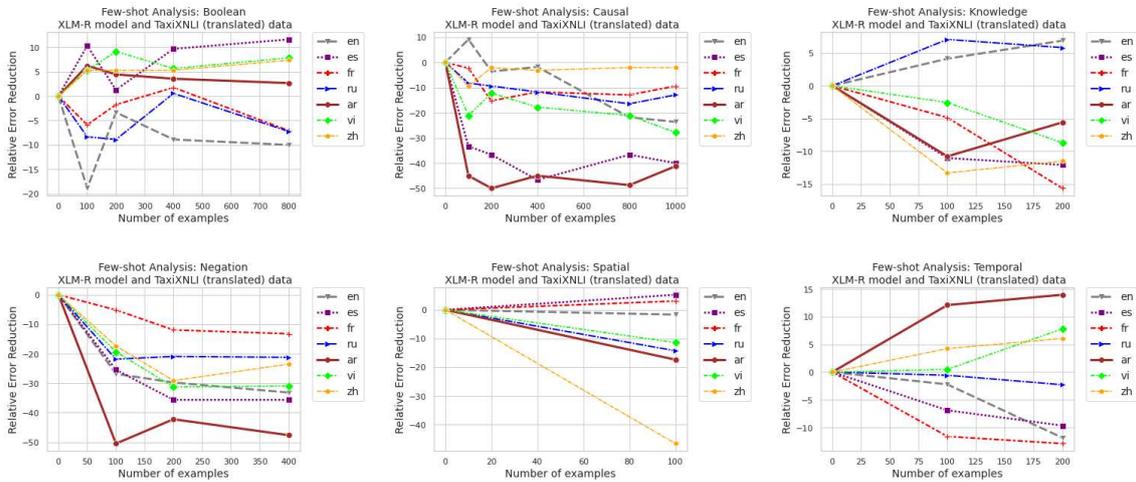

Figure 5: **Few Shot Accuracy (TAXIXNLI (diagnostic), XLM-R):** For each TAXINLI category and language, we plot Accuracy on TAXIXNLI (diagnostic) test, when XLM-R MultiNLI model is fine-tuned on $N$ examples in that language and category

and 2, we observe that the transferability of different inference types are non-uniform, especially in the case of XLM-R. In particular, transferability of `negation` (except for Swahili) is a lot better than `lexical`, `syntactic`, `boolean`, or `knowledge` inference types. The results on the TaxiXNLI (diagnostic) test confirms these observa-tions. Further, from the results on TaxiXNLI (translated), in Table 5, we observe that the transfer gap for some categories such as `negation`, `boolean` can be reduced significantly by increasing the monolingual pre-training data (from M-BERT to XLM-R), whereas for categories such as `knowledge` and `deductions` the transfer gap is not reduced

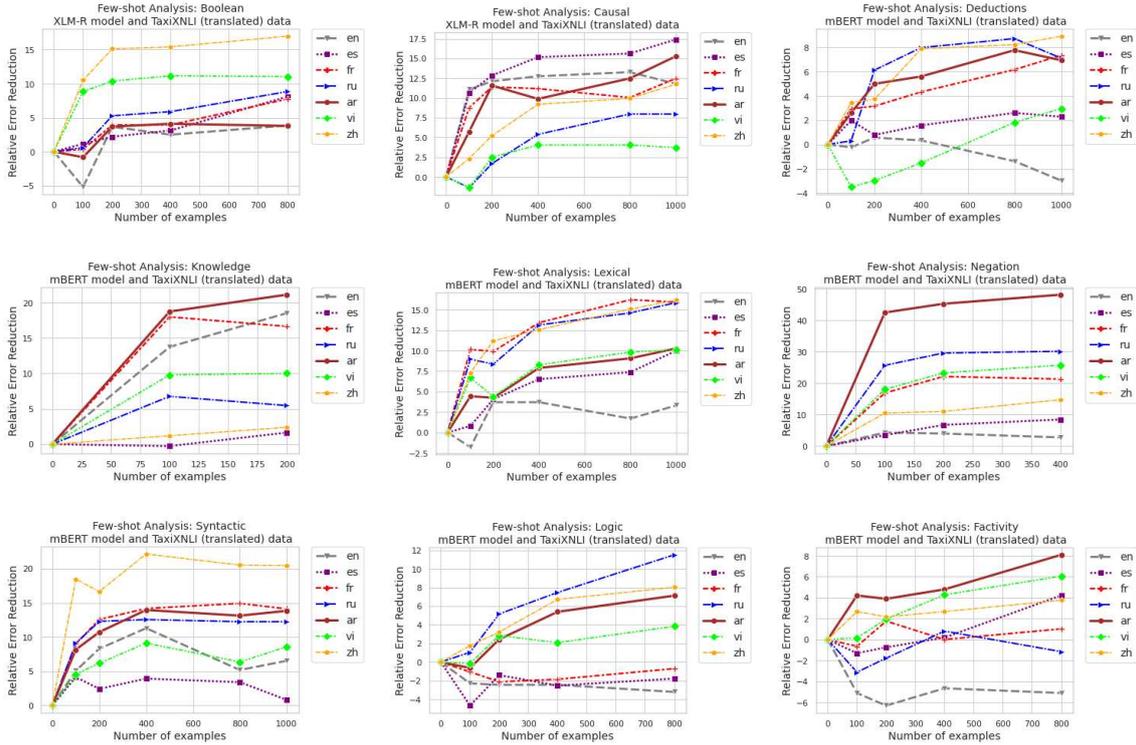

Figure 6: **Few Shot Accuracy (TaxiXNLI (translated), mBERT):** For each TaxiNLI category and language, we plot Accuracy on TaxiXNLI (translated) test, when mBERT MultiNLI model is fine-tuned on $N$ examples in that language and category

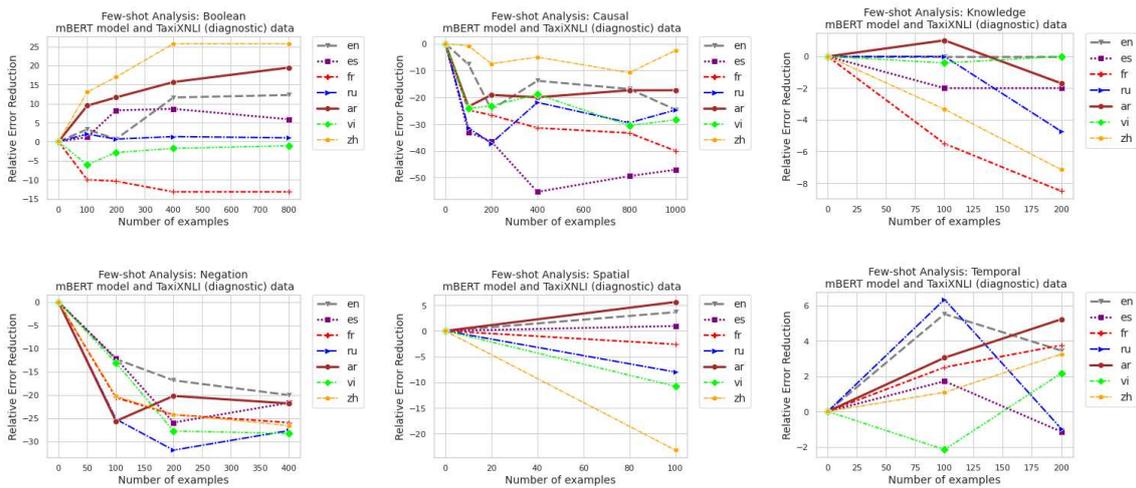

Figure 7: **Few Shot Accuracy (TaxiXNLI (diagnostic), mBERT):** For each TaxiNLI category and language, we plot Accuracy on TaxiXNLI (diagnostic) test, when mBERT MultiNLI model is fine-tuned on $N$ examples in that language and category

much.

**Why does zero-shot transfer of Negation seem easy? And, why is Swahili the odd-one-out?** Negation seems easiest to transfer. A possible cause is in most languages, there are standard negation keywords. In English (among 1376 examples, TaxiXNLI-translated), most frequently occurring words are: not (543), never (172), no (260), didn't (73), No (42), nothing (45). It's possible that pretrained LMs learn correspondence between English and target language negation keywords. However, in Swahili, zero-shot transfer for Negation is bad. We observe that Swahili uses fused morphemes to indicate negation; for example, negation marker "Si" used for the pronoun "Mimi" (I), "Hatu" for "Sisi" (we). Sentence " I am not singing" becomes "Siimbi"; "We are not singing" becomes "Hatuiimbi". It's possible that with the comparatively limited pre-training data, XLM-R does not learn the correspondence between separate negative particles and fused morphemes.

**Language features or Reasoning categories – which affects transfer more?** Inspired by Lauscher et al. (2020), we extend their regression analysis to include category features. Since, multiple categories can exist per example, we create a feature-set for each example in TaxiXNLI (translated) test set. We include Lang2Vec features of its corresponding language (syntax, phonology, inventory, family, and GEO), and one-hot category vector (1 if it belongs to the category otherwise zero). We then use logistic regression (LR) and linear discriminant analysis (LDA) to predict which factors influence XLM-R's correct prediction (0/1) on TaxiXNLI (translated) dataset. As shown in Figure 3, with high $p$-values, `negation`, `logic` and `deductions` affect XLM-R's zero-shot performance. Interestingly, for mBERT, the language features correlate more highly with correct prediction.

**Do all categories improve under few-shot?** We plot the relative error reduction in the few-shot scenario (refer Fig. 4, and Fig. 5) for XLM-R. We only retain the high-resource languages for both, as translations for high-resource languages are quite accurate. In the TaxiXNLI (translated) set, for `boolean`, `causal`, `deduction` and `knowledge`, all languages improve uniformly with small variations. However, on TaxiXNLI (diagnostic) test, there is a large decrease in RER for `causal` and other categories. To investigate deeper, we analyze if few-shot improves crosslingual alignment. For each category, we take (`[CLS]`) embeddings of the English (pivot) sentence and its translated counterpart in target language from TaxiXNLI (translated) — and compute their pair-wise distances. In Table 6, We report the difference between average distance after few-shot (highest N) and at zero-shot. Both $L^2$ distance (and cosine) measure shows that for `negation`, the crosslingual alignment is increasing after few-shot. Though, results on the diagnostic set calls for careful experiments to test generalization.

|  | ar | es | fr | hi | ru | sw | ur | vi | zh |
|---|---|---|---|---|---|---|---|---|---|
| **Negation** | -0.73 | -0.23 | 0.06 | -0.02 | 0.005 | -1.3 | -0.64 | -0.46 | -0.55 |
| **Min** | -0.912 | -0.55 | -0.797 | -0.75 | -0.54 | -1.3 | -1.009 | -0.74 | -0.88 |
| **Max** | 0.162 | 0.96 | 0.39 | 0.56 | 0.16 | -0.03 | -0.01 | 0.27 | 0.35 |

Table 6: Difference between L2 distances of sentence-pairs for each categories before and after few-shot. For *negation*, there is large decrease in distances, indicating higher crosslingual alignment.

## 5 Findings and Future Work

We proposed a multilingual extension of a reasoning type-annotated NLI dataset (TaxiNLI) in English – TaxiXNLI (translated) and a gold test set TaxiXNLI (diagnostic) to analyze transfer efficiency of multilingual language models. Our analysis strongly indicates that reasoning types play an important role in transfer capability. Our few-shot results demonstrate that for types such as Negation, improvement in transfer accuracy co-occurs with improved crosslingual alignment. For causal, few-shot models fail to generalize well on the diagnostic set; and some categories hardly improve. In summary, NLI being central to NLU and NLP, this dataset and our ensuing analysis provides a way to broadly quantify the gaps in zero-shot transfer across various linguistic and logical dimensions.

# Appendix: Analyzing the Effects of Reasoning Types on Crosslingual Transfer Performance

**Karthikeyan K**
Duke University
karthikeyan.k@duke.edu

**Aalok Sathe**
Massachusetts Institute of Technology
asathe@mit.edu

**Somak Aditya** and **Monojit Choudhury**
Microsoft Research India
{t-soadit,monojitc}@microsoft.com

## 1 Zero Shot Crosslingual Transfer

Similar to the zero-shot analysis for XLM-R, we include the zero-shot transfer gap plots for each category and language pair on both datasets in Figure 1. Interestingly, even for mBERT, transfer for `negation` seems easier in both datasets; and Swahili remains the outlier.

### 1.1 Confounder Analysis for mBERT

Similar to the analysis done for TaxiXNLI (translation), we plot the coefficients obtains from Logistic Regression and Linear Discriminant Analysis for both XLM-R and mBERT in Figure 2.

## 2 Few Shot Cross-Lingual Transfer

In this section, we present both Relative Error Reduction and accuracy when XLM-R and M-BERT are fine-tuned on a few examples of TaxiXNLI training data for a category-language pair. For XLM-R, we report our results on TaxiXNLI (Translated) and TaxiXNLI (diagnostic) on Figures 3 and 4 respectively. We exclude the mBERT results on few-shot to abide by space limitations, where we make similar observations around improvement patterns, and even the lack of generalization for `causal`. We also report the same for mBERT in Figures 5 and 6.

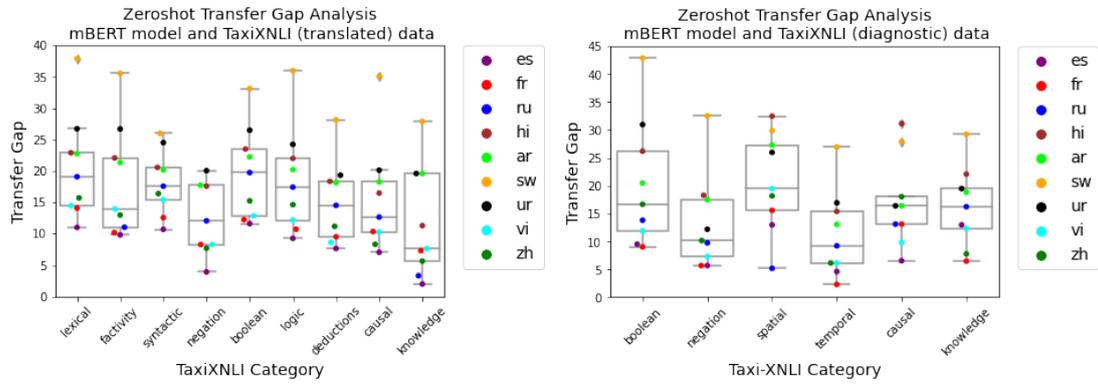

Figure 1: For each TaxiNLI category and language, we plot the transfer gap on mBERT, for both datasets.

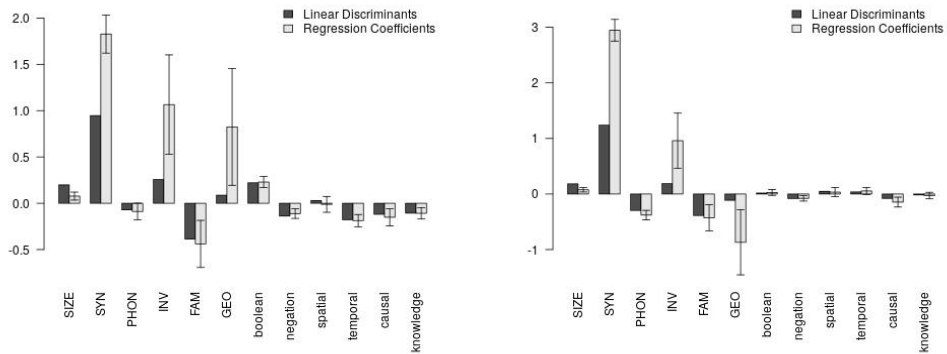

Figure 2: LDA and regression Coefficients obtained to model the correctness of NLI prediction by XLM-R(left) and mBERT (right), given the language features and reasoning categories on TaxiXNLI (diagnostic) dataset

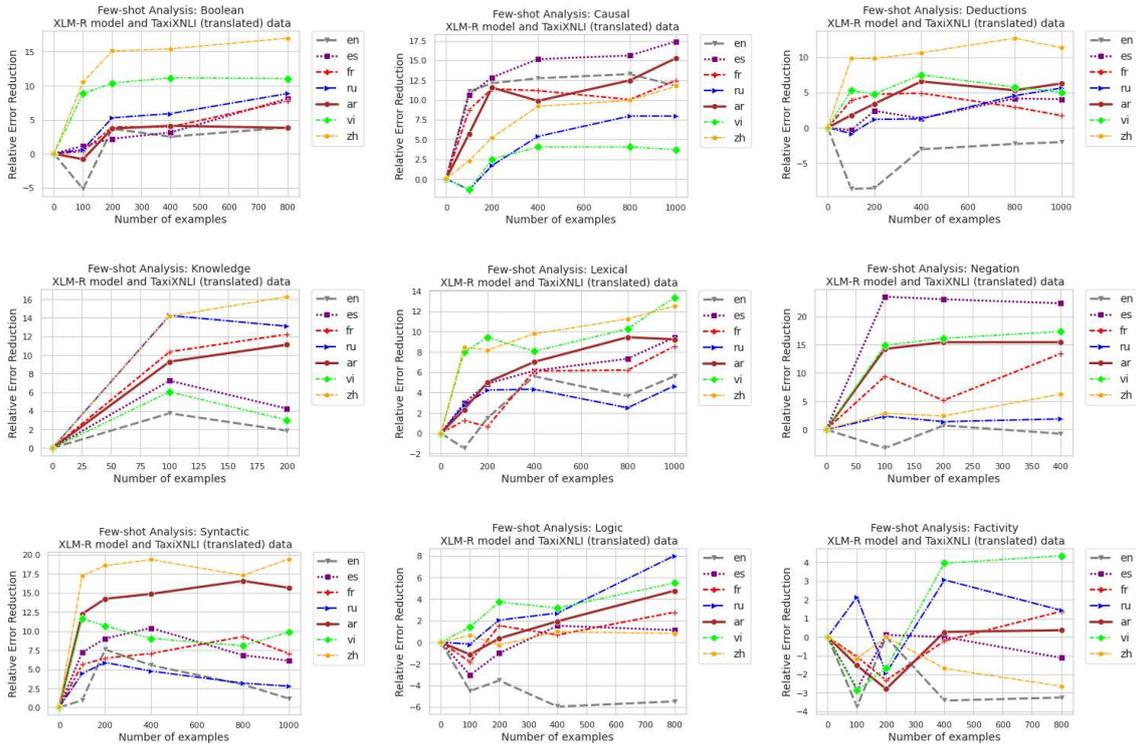

Figure 3: **Few Shot Accuracy (TaxiXNLI (translated), XLM-R):** For each TaxiNLI category and language, we plot Accuracy on TaxiXNLI (translated) test, when XLM-R MultiNLI model is fine-tuned on $N$ examples in that language and category

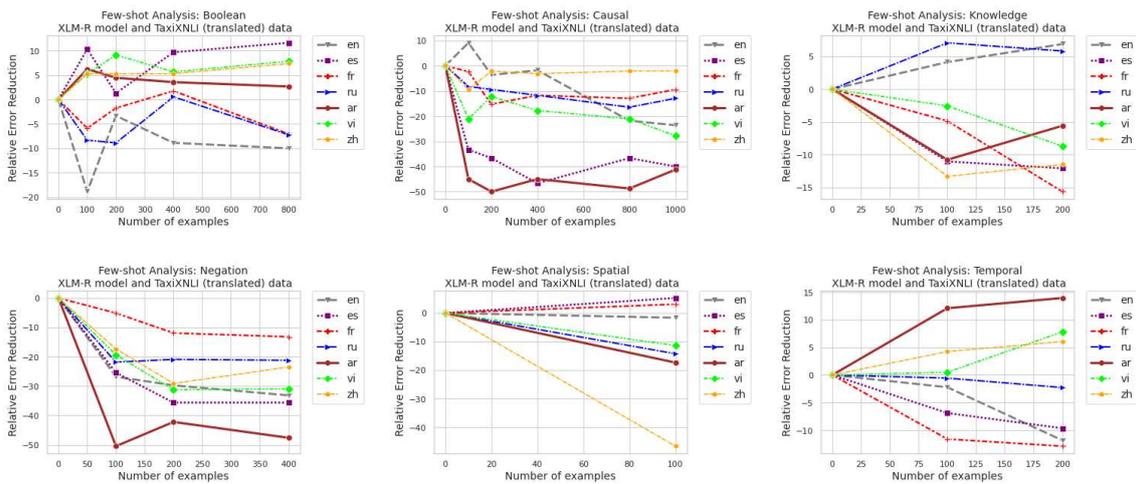

Figure 4: **Few Shot Accuracy (TaxiXNLI (diagnostic), XLM-R):** For each TaxiNLI category and language, we plot Accuracy on TaxiXNLI (diagnostic) test, when XLM-R MultiNLI model is fine-tuned on $N$ examples in that language and category

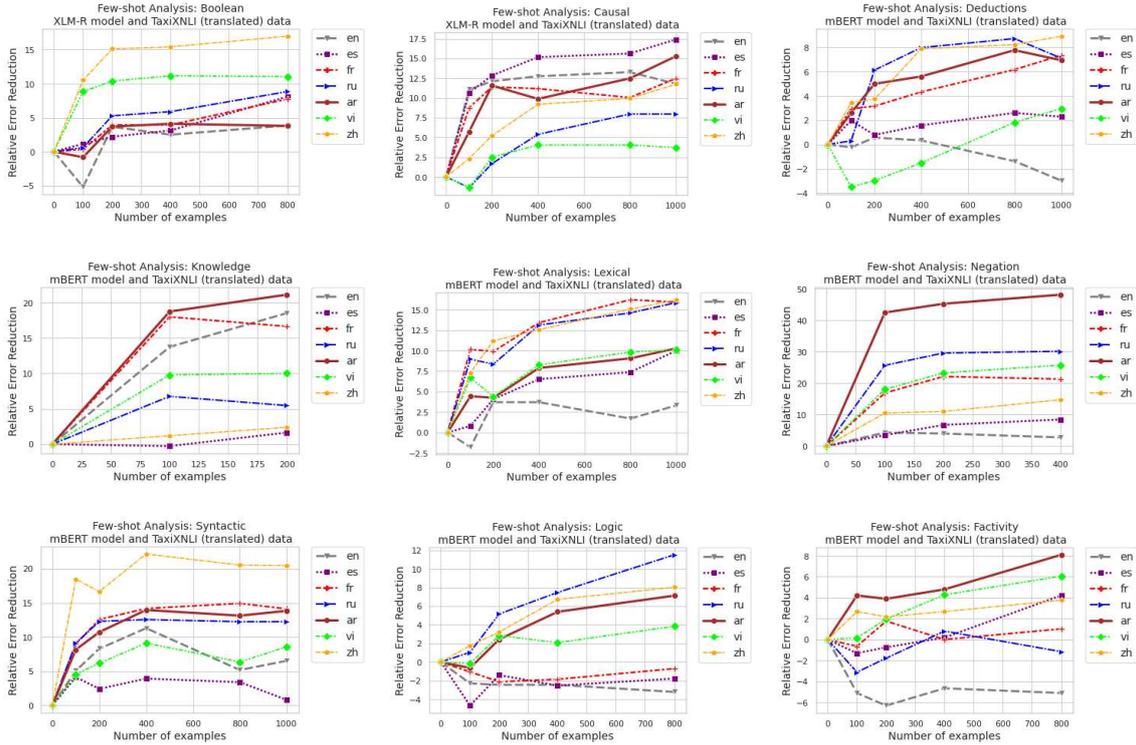

Figure 5: **Few Shot Accuracy (TAXIXNLI (translated), mBERT):** For each TAXINLI category and language, we plot Accuracy on TAXIXNLI (translated) test, when mBERT MultiNLI model is fine-tuned on $N$ examples in that language and category

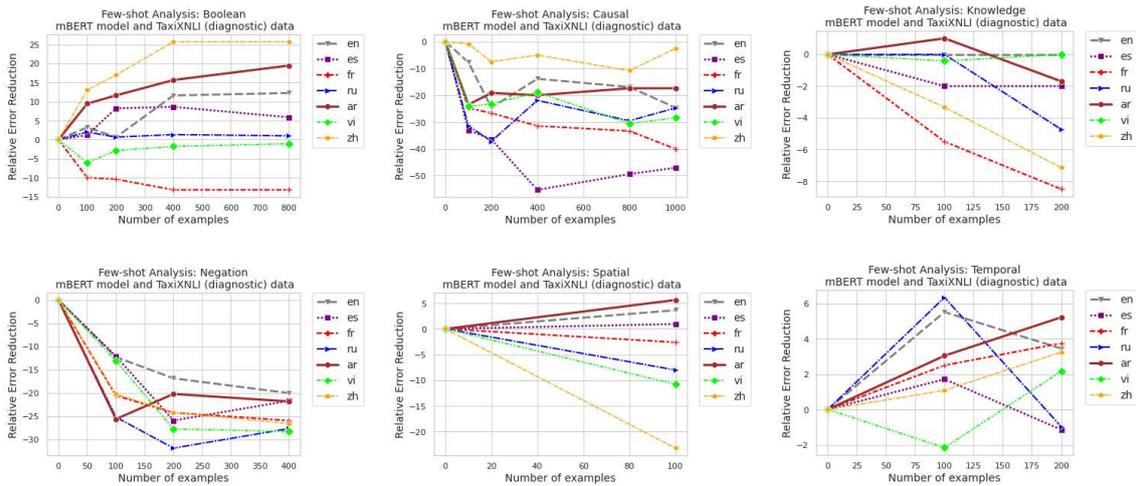

Figure 6: **Few Shot Accuracy (TAXIXNLI (diagnostic), mBERT):** For each TAXINLI category and language, we plot Accuracy on TAXIXNLI (diagnostic) test, when mBERT MultiNLI model is fine-tuned on $N$ examples in that language and category